\documentclass[runningheads]{llncs}

\usepackage{eccv}

\usepackage{eccvabbrv}
\usepackage{graphicx}
\usepackage{lipsum}
\usepackage{algorithm}
\usepackage{algpseudocode}
\usepackage{dsfont}
\usepackage{multirow}
\usepackage{enumitem}
\usepackage{booktabs}
\usepackage{wrapfig}
\usepackage[accsupp]{axessibility}
\usepackage[breaklinks,colorlinks,citecolor=eccvblue]{hyperref}
\usepackage{orcidlink}

\newcommand{\corrauth}{\unskip$^\star$}

\begin{document}

\title{Progressive Learned Image Compression for Machine Perception} 
\titlerunning{Progressive Learned Image Compression for Machine Perception}

\author{
  Jungwoo Kim\inst{1,2}\orcidlink{0009-0008-1472-3927} \and
  Jun-Hyuk Kim\inst{3}\thanks{Corresponding authors}\orcidlink{0000-0002-7980-4583} \and
  Jong-Seok Lee\inst{1,2}\corrauth\orcidlink{0000-0002-8038-1119}
}

\authorrunning{J. Kim et al.}

\institute{
  School of Integrated Technology, Yonsei University 
  \and
  BK21 Graduate Program in Intelligent Semiconductor Technology, Yonsei University
  \and
  Department of Artificial Intelligence, Chung-Ang University \\
\email{\{kjungwoo, jong-seok.lee\}@yonsei.ac.kr} \\
\email{junhyukkim@cau.ac.kr}
}

\maketitle

\begin{abstract}
  Recent advances in learned image codecs have extended from human perception toward machine perception.
  However, progressive image compression with fine granular scalability (FGS)--which enables decoding a single bitstream at multiple quality levels--remains unexplored for machine-oriented codecs.
  In this work, we propose \textbf{PICM-Net}, a progressive learned image compression codec for machine perception built on trit-plane coding.
  Starting from a human-oriented codec, we adapt it for machine perception via a spatial-frequency modulation adapter (SFMA), a hyper-synthesis low-rank adapter (HSLoRA), and progressive decoding-aware training, and analyze how task-driven adaptation alters symbol prioritization for progressive transmission.
  To further support real-world deployment, we introduce an \textbf{adaptive decoding controller} that dynamically determines the necessary decoding level at inference time, requesting additional bits only when the current suitability is insufficient for the desired confidence level.
  Extensive experiments demonstrate that PICM-Net achieves efficient and adaptive progressive transmission while maintaining strong downstream classification performance.
  Our code is available at \url{https://github.com/kjungwoo03/PICM-Net}.
  \keywords{Learned Image Compression, Semantic Scalability, Progressive Compression}
\end{abstract}


\section{Introduction}
\label{sec:intro}

Traditional image codecs, such as JPEG~\cite{wallace1992jpeg}, JPEG2000~\cite{skodras2001jpeg2000}, WebP~\cite{webp2017}, and VVC~\cite{bross2021overview}, have been primarily designed to optimize visual quality for human perception.
Recently, deep learning-based learned image compression methods~\cite{balle2018variational, minnen2018joint, cheng2020learned, kim2020efficient, zou2022devil, he2022elic, liu2023learned, kim2022joint, yu2023lut, zeng2025mambaic} have achieved superior rate-distortion (RD) performance compared to traditional codecs through end-to-end optimization with deep neural networks.
However, their optimization objectives remain centered on human visual fidelity, typically using perceptual loss functions and metrics such as MSE, MS-SSIM~\cite{wang2003multiscale} or LPIPS~\cite{zhang2018unreasonable}.

Recently, with the rapid growth of machine vision applications--such as autonomous driving~\cite{song2025don, lee2026iros}, surveillance systems~\cite{cho2022part, yuan2024towards, hu2025enhanced}, and remote sensing~\cite{liu2025efficient, yun2024powdew}--images are increasingly consumed by machines rather than humans.
This paradigm shift has motivated the development of machine-oriented codecs~\cite{yan2024taskoriented, guo2024unified, feng2022image, liu2023icmhnet, le2021image, chen2023transtic, park2025test, liu2024rate, chen2025unirestore, li2024image, tatsumi2025explicit, zhang2024all, shindo2024image, lee2025slim, shindo2025guided}, which prioritize task performance over human perceptual quality by optimizing for downstream vision tasks, such as classification~\cite{he2016deep, dosovitskiy2021an, russakovsky2015imagenet}, detection~\cite{ren2015faster, redmon2016yolo, carion2020detr}, and segmentation~\cite{long2015fcn, chen2018deeplab, zhou2018unetpp}.
These approaches have demonstrated that task-driven image compression can achieve better performance at lower bitrates by focusing on semantically important features rather than pixel-level reconstruction quality.
Earlier works~\cite{feng2022image, le2021image} primarily focused on end-to-end optimization for machine-oriented codecs, while recent works~\cite{liu2024rate, li2024image, zhang2024all, chen2023transtic, park2025test} have adopted approaches to fine-tune existing human-oriented codecs to adapt them for multiple machine vision tasks.

Meanwhile, in the field of human-oriented compression, progressive image coding--also known as fine granular scalability (FGS)~\cite{shapiro1993embedded, taubman2000high, schaar2001mpeg}--has been studied to enable multi-stage decoding and adaptive bit transmission~\cite{presta2025efficient, guo2025oscar, lee2022dpict, jeon2023context, hojjat2023progdtd, lee2024deephq, yang2025progressive, li2025onceforall, lu2021progressive}.
This paradigm allows a single bitstream to be decoded at various quality levels, providing early access to coarse reconstructions and improving efficiency in real-world scenarios where network bandwidth fluctuates~\cite{stockhammer2011das, ohm2005scalable}.
Despite its effectiveness for human perception and real-world settings, progressive image compression has not yet been explored for machine-oriented codecs.
This gap motivates us to revisit progressive compression from a machine perspective, aiming to develop a codec that combines the flexibility of progressive image compression with the task-aware optimization of machine-oriented compression.

\begin{figure*}[t]
  \centering
  \includegraphics[width=0.9\textwidth]{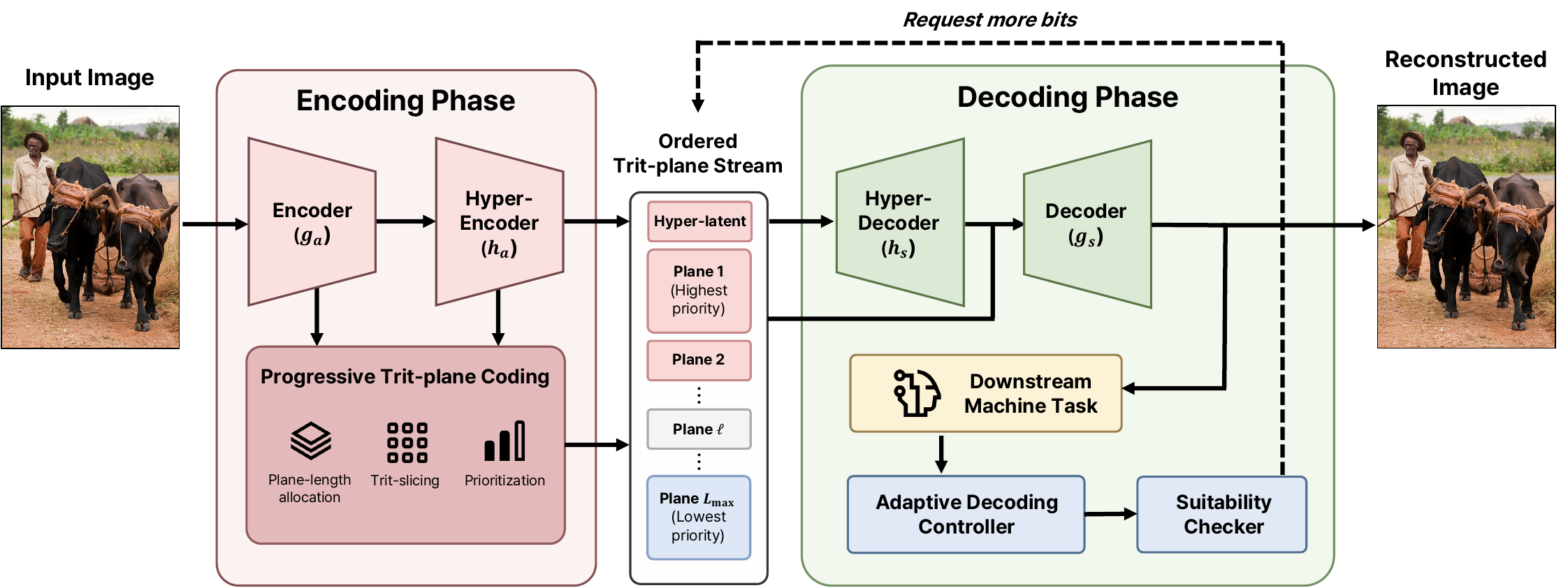}
  \caption{
    Overview of our proposed codec, PICM-Net. The encoder produces an ordered trit-plane bitstream via plane-length allocation, trit-slicing, and prioritization. At inference, the adaptive decoding controller evaluates downstream task suitability at each decoding level, requesting additional bits only when the confidence threshold is not satisfied.}
    \vspace{-1.8em}
  \label{fig:overview} 
\end{figure*}

In this work, we propose \textbf{PICM-Net, to the best of our knowledge the first progressive image codec for machine perception}.
Our codec builds on three components: (i) progressive trit-plane coding that decomposes latent representations into ternary digits (trits) for coarse-to-fine transmission, (ii) a spatial-frequency modulation adapter (SFMA)~\cite{li2024image} to adapt the codec toward machine perception, and (iii) an adaptive decoding controller that dynamically selects the decoding level based on downstream prediction confidence (see Fig.~\ref{fig:overview}).

Starting from the human-oriented TIC codec~\cite{lu2022tic}, we adapt it for machine perception while preserving the progressive trit-plane coding scheme.
Concretely, we extend SFMA to the hyperprior-based TIC architecture and introduce a lightweight LoRA module~\cite{hu2022lora} on the hyper-synthesis networks (HSLoRA), to align entropy parameters with the adapted latent distribution.
We further apply a progressive decoding-aware training objective~\cite{hojjat2023progdtd} that supervises downstream tasks not only at full reconstruction but also at randomly sampled trit-plane prefixes, encouraging early trits to carry discriminative semantic content.

A key distinction between human and machine perception lies in their quality requirements.
For humans, higher reconstruction quality is usually preferable--more bits consistently yield better perceptual quality.
Machines, however, are indifferent to quality beyond what is sufficient for reliable task inference: once a classifier can correctly identify an image, additional bits provide no benefit.
This asymmetry motivates our \textbf{adaptive decoding controller}, which monitors classifier logits at each decoding stage and requests additional bits only when the estimated \emph{suitability} (estimated correctness) is below a desired confidence level--enabling efficient and adaptive bitstream transmission without committing to a fixed decoding level.

Extensive experiments demonstrate that PICM-Net achieves superior task performance and transmission efficiency compared to existing progressive and machine-oriented codecs, while uniquely supporting fine-grained scalability for machine vision.
Our contributions are as follows:
\begin{itemize}[leftmargin=*,label=\textbf{·}]
  \item We propose \textbf{PICM-Net}, the first progressive image compression codec for machine perception, built on trit-plane coding with machine-aware prioritization.
  \item We introduce \textbf{machine-oriented progressive adaptation} combining SFMA and HSLoRA with progressive decoding-aware training, which aligns the entropy model with adapted latent statistics while ensuring task performance across all decoding levels.
  \item We introduce an \textbf{adaptive decoding controller}, which dynamically determines the necessary decoding level at inference time based on predicted suitability--a critical capability for real-world deployment under bandwidth constraints.
  \item We provide extensive experiments validating the effectiveness of our approach against existing progressive codecs and machine-oriented codecs.
\end{itemize}
\section{Related Work}
\label{sec:rework}

\subsection{Learned Image Compression}
\label{sec:rework-lic}
Deep learning-based learned image compression (LIC) has achieved remarkable rate-distortion efficiency, surpassing traditional codecs through end-to-end optimization.
Standard LIC codecs follow an autoencoder architecture: the encoder maps images to quantized latent representations, which are entropy-coded and decoded to reconstruct images.
Architectures have evolved from convolutional networks~\cite{balle2018variational, minnen2018joint, cheng2020learned} to transformers~\cite{lu2022tic, liu2023learned, kim2022joint, zeng2025mambaic}, while entropy models have advanced from factorized priors~\cite{balle2018variational} to spatial and channel-wise autoregressive models~\cite{minnen2018joint, he2022elic, cheng2020learned}, enabling increasingly accurate probability estimation.

\subsection{Image Compression for Machine}
\label{sec:rework-icm}
With the growing demands of machine vision applications, there has been increasing interest in compression methods optimized for downstream task performance rather than human perception.
Early works~\cite{le2021image, feng2022image} jointly optimized the codec and task model end-to-end, achieving better rate-accuracy trade-offs by preserving task-relevant features, but requiring separate training for each task.
Multi-task approaches~\cite{yan2024taskoriented, guo2024unified} addressed this with shared encoders and task-specific decoders under scalable coding, yet still demanded full system training from scratch.
More recently, fine-tuning-based methods~\cite{liu2023icmhnet, chen2023transtic, li2024image, zhang2024all, park2025test} have adapted pretrained human-oriented codecs~\cite{he2022elic, cheng2020learned} using lightweight modules such as LoRA, achieving flexible task-aware compression at lower training cost.
Despite their effectiveness, all existing machine-oriented codecs remain limited to a single, non-progressive encoding pass.

\subsection{Progressive Image Compression}
\label{sec:rework-pic}
Progressive image compression enables coarse-to-fine reconstruction from a single bitstream, allowing partial decoding at multiple quality levels as more bits are received.
Early approaches~\cite{lu2021progressive, hojjat2023progdtd} explored scaling and rounding of latent representations to support multi-level decoding.
More recent learned codecs~\cite{lee2022dpict, jeon2023context, lee2024deephq, presta2025efficient} adopt multi-slice latent structures for sequential refinement, with some methods leveraging diffusion-based generative models~\cite{yang2025progressive, li2025onceforall} for higher-quality reconstruction at low bitrates.
However, all existing progressive methods target human perceptual quality, leaving machine-aware progressive compression unexplored.

\section{Methods}
\label{sec:methods}

\subsection{Framework Overview}
\label{sec:methods-framework-overview}

\begin{wrapfigure}[9]{r}{0.35\textwidth}
    \centering
    \vspace{-2.0em}
    \includegraphics[width=\linewidth]{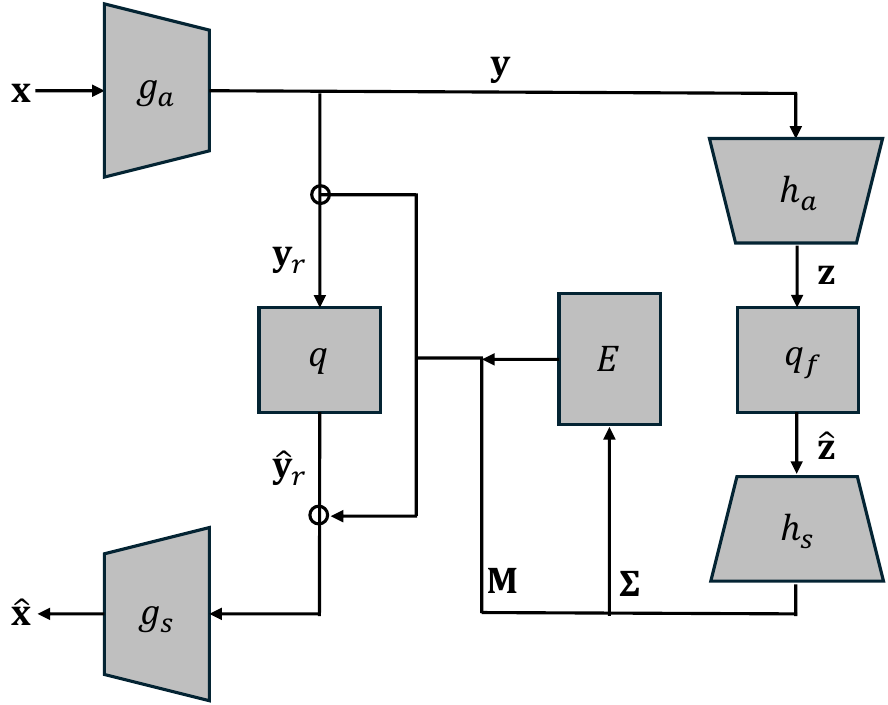}
    \caption{Overall architecture of our proposed codec.}
    \label{fig:p2-scheme}
    \vspace{-1.5em}
\end{wrapfigure}

We build upon a standard hyperprior-based autoencoder architecture~\cite{balle2018variational, minnen2018joint, cheng2020learned} for learned image compression (see Fig.~\ref{fig:p2-scheme}).
The encoder $g_a$ maps an input image $\mathbf{x}\in\mathbb{R}^{H\times W\times 3}$ to a latent representation $\mathbf{y}\in\mathbb{R}^{H/16\times W/16\times C}$, which is further compressed by the hyperprior encoder $h_a$ into a hyperlatent $\mathbf{z}\in\mathbb{R}^{H/64\times W/64\times C}$.
The hyperprior decoder $h_s$ then decodes the quantized hyperlatent $\mathbf{\hat{z}}$ 
to produce the mean $\mathbf{M}$ and scale $\mathbf{\Sigma}$ parameters of the latent distribution.

To enable progressive transmission, we apply residual quantization: the latent is centered as $\mathbf{y}_r = \mathbf{y} - \mathbf{M}$ and uniformly quantized as $\mathbf{\hat{y}}_r = q(\mathbf{y}_r)$, where $q(\cdot)$ denotes rounding.
Rather than encoding $\mathbf{\hat{y}}_r$ directly, we decompose it into trit-planes for coarse-to-fine progressive coding.
On the decoder side, the reconstructed latent is recovered as $\mathbf{\hat{y}} = \mathbf{\hat{y}}_r + \mathbf{M}$, and passed through the synthesis network $g_s$ to reconstruct $\mathbf{\hat{x}}$.

Our framework integrates three components on top of this backbone: (i) progressive trit-plane coding for multi-level bitstream decoding (Sec.~\ref{sec:methods-trit-plane}), (ii) machine-oriented adaptation via lightweight modules to fine-tune the codec toward downstream task performance (Sec.~\ref{sec:method-adaptation}), and (iii) an adaptive decoding controller that dynamically selects the decoding level at inference time based on predicted suitability (Sec.~\ref{sec:methods-adaptive-decoding-controller}).

\subsection{Progressive Trit-plane Coding}
\label{sec:methods-trit-plane}

We adopt trit-plane coding~\cite{lee2022dpict, jeon2023context} to enable progressive transmission from a single bitstream.
Each coefficient is decomposed into ternary digits (trits) and encoded plane-by-plane, yielding a coarse-to-fine bitstream that can be decoded at multiple intermediate quality levels.

\vspace{0.5em}
\noindent \textbf{Plane-length allocation.}
The number of trit-planes required per coefficient is determined by its predicted 
scale parameter $\hat{\sigma}$, which characterizes the width of the Gaussian prior.
We define an effective range as:
\begin{equation}
  \text{tail}_c = 2\kappa\hat{\sigma}_c,
\end{equation}
where $\kappa = -{\Phi}^{-1}(10^{-9}/2)$ is a constant from the inverse CDF of the standard Gaussian, capturing the tail probability at $10^{-9}$.
The required number of ternary digits for coefficient $c$ is then:
\begin{equation}
  L_c = \max\left(1,\, \lceil \log_3(\text{tail}_c) \rceil\right),
\end{equation}
where the $\max(1, \cdot)$ ensures every coefficient is allocated at least one trit-plane, even when $\hat{\sigma}_c$ is near zero.
Coefficients with larger $\hat{\sigma}_c$ (wider distributions) are allocated more planes, while those with smaller $\hat{\sigma}_c$ need fewer.
The global maximum $L_{\text{max}} = \max_c L_c$ defines the total number of planes.

\vspace{0.5em}
\noindent \textbf{Ternary decomposition.}
Each quantized coefficient $\hat{y}_c$ is mapped to a non-negative integer 
$s_c = \hat{y}_c + \lfloor 3^{L_c}/2 \rfloor$ and decomposed into its ternary representation:
\begin{equation}
  s_c = \sum_{\ell=1}^{L_c} d_{c,\ell} \cdot 3^{L_c - \ell}, \quad d_{c,\ell} \in \{0,1,2\},
\end{equation}
where $d_{c,\ell}$ is the $\ell$-th trit of coefficient $c$.
Stacking across all coefficients yields a trit tensor 
$\mathbf{N} \in \mathbb{Z}^{S \times L_{\text{max}}}$, where $S = (H/16)\cdot(W/16)\cdot C$.

\vspace{0.5em}
\noindent \textbf{Plane-wise entropy coding.}
For each plane $\ell$, we construct conditional probability models from the Gaussian scale $\mathbf{\Sigma}$.
Specifically, for a coefficient with $L_c$ planes, we precompute a PMF over $3^\ell$ symbols by integrating $\mathcal{N}(0, \hat{\sigma}_c^2)$ over integer bins, then marginalize over previously decoded planes to obtain refined conditional distributions.
Each successive plane narrows the interval of possible values, progressively reducing uncertainty in $\hat{y}_c$.
The decoder mirrors this process, reconstructing each coefficient from the decoded trits in order.

\vspace{0.5em}
\noindent \textbf{Symbol prioritization.}
Within each plane, the transmission order of symbols affects rate-distortion efficiency.
Prior works propose two main strategies: expected variance-based ordering~\cite{lee2022dpict,jeon2023context} prioritizes symbols by their expected MSE reduction per bit, computing conditional variance reductions across the Gaussian prior, while sigma-based ordering~\cite{presta2025efficient, lu2021progressive} instead orders coefficients directly by their scale parameter $\hat{\sigma}_c$, offering a computationally efficient proxy for the same objective.
We adopt sigma-based ordering as our default prioritization strategy, and empirically show in Sec.~\ref{sec:ablation} that it is more efficient than expected variance-based ordering.

\begin{figure*}[t]
  \centering
  \begin{subfigure}[t]{0.6\textwidth}
    \centering
    \includegraphics[width=\linewidth]{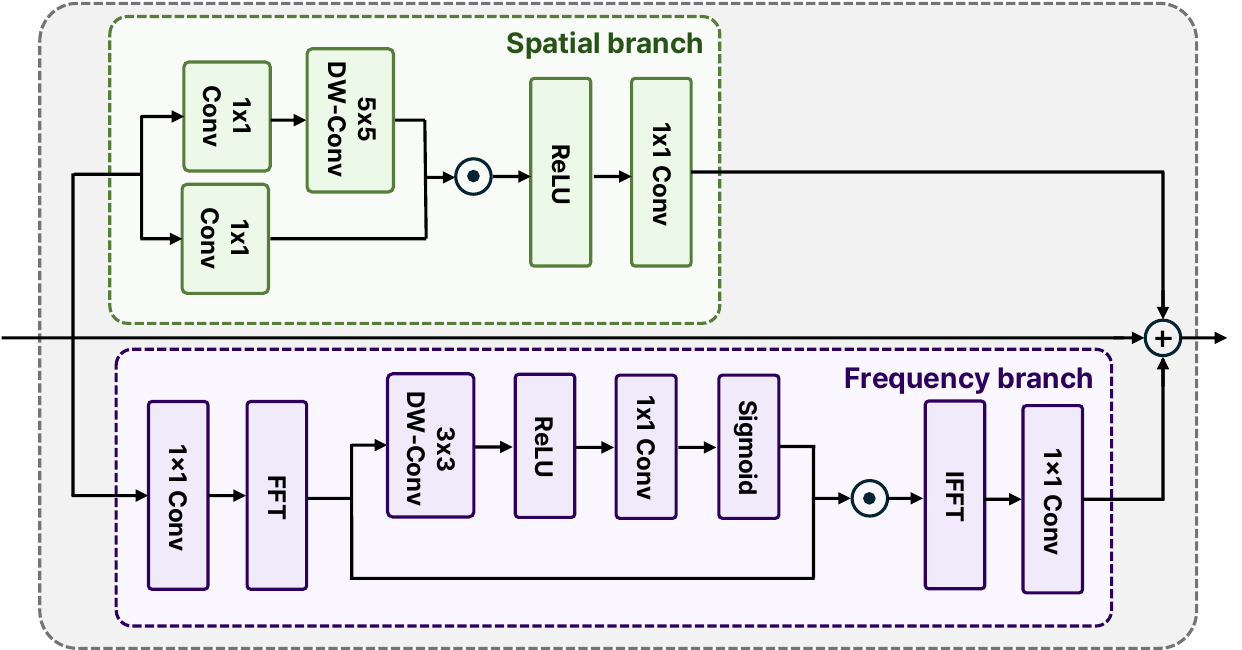}
    \caption{SFMA Module}
    \label{fig:sfma-module}
  \end{subfigure}
  \hspace{0.5em}
  \begin{subfigure}[t]{0.28\textwidth}
    \centering
    \includegraphics[width=\linewidth]{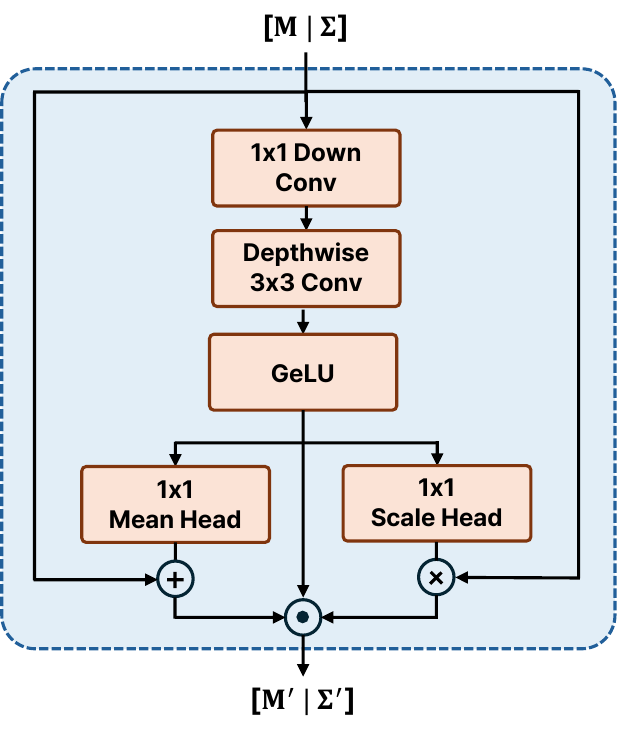}
    \caption{HSLoRA Module}
    \label{fig:hslora-module}
  \end{subfigure}
  \vspace{-0.5em}
  \caption{
    Architecture of spatial-frequency modulation adapter (SFMA) and hyper-synthesis low-rank adapter (HSLoRA).
  }
  \vspace{-1.5em}
  \label{fig:adaptation}
\end{figure*}

\subsection{Machine-Oriented Progressive Adaptation}
\label{sec:method-adaptation}
Although trit-plane coding enables scalable decoding, the human-oriented TIC backbone~\cite{lu2022tic} is originally trained to prioritize visual reconstruction quality rather than downstream task performance.
We adapt it toward machine perception by freezing the pretrained backbone and training only lightweight adaptation modules.

\vspace{0.5em}
\noindent \textbf{Adaptation modules.}
Our backbone consists of two encoder-decoder pairs, the main codec ($g_a$, $g_s$) and the hyperprior ($h_a$, $h_s$), and we adapt each with a lightweight module suited to its role (see Fig.~\ref{fig:adaptation}).
First, we insert spatial-frequency modulation adapters (SFMA)~\cite{li2024image} into intermediate features of both the encoder $g_a$ and decoder $g_s$ (Fig.~\ref{fig:sfma-module}).
Each SFMA module modulates the feature maps in both spatial and frequency domains, allowing the codec to suppress visually important but task-irrelevant information while amplifying semantically discriminative features--without modifying the backbone weights.

For the hyperprior pair, we introduce HSLoRA, a $1{\times}1$ convolutional low-rank adapter (Fig.~\ref{fig:hslora-module}), applied only to the hyper-synthesis network $h_s$. 
We leave the hyper-analysis network $h_a$ unadapted because its output $\textbf{z}$ is entropy-coded with a factorized density model~\cite{balle2018variational} fixed at backbone pre-training--adapting $h_a$ would shift the distribution of $\textbf{z}$ and decalibrate the fixed model, which cannot be refit under our frozen-backbone design. 
The hyper-synthesis network $h_s$, by contrast, directly outputs the mean $\textbf{M}$ and scale $\mathbf{\Sigma}$ of the conditional Gaussian used to code $\textbf{y}$, so adapting $h_s$ alone suffices to realign this distribution with the latent statistics shifted by SFMA, without disturbing the fixed entropy model governing $\textbf{z}$.

\vspace{0.5em}
\noindent \textbf{Progressive decoding-aware training.}
We optimize the adaptation modules with a machine-oriented rate-distortion objective:
\begin{equation}
  \mathcal{L}_{\mathrm{machine}} = R + \lambda_{\mathrm{task}}
  \left(\mathcal{L}_{\mathrm{CE}}(f(\hat{\textbf{x}}), t) + \lambda_{\mathrm{mse}}\|\textbf{x} - \hat{\textbf{x}}\|_2^2\right),
  \label{eq:loss-function}
\end{equation}
where $R$ is the bitrate, $t$ denotes the ground-truth class label, $f(\cdot)$ is a pretrained downstream classifier, and the MSE term serves as a regularizer for stable training.

Optimizing only at full reconstruction, however, does not guarantee task performance at intermediate decoding levels.
To address this, we apply progressive decoding-aware training~\cite{hojjat2023progdtd}, which extends Eq.~\ref{eq:loss-function} by sampling the decoding level at each training iteration:
\begin{equation}
  \mathcal{L}_{\mathrm{prog}} = \mathbb{E}_{\ell}
  \left[ R^{(\ell)} + \lambda_{\mathrm{task}}
  \left(\mathcal{L}_{\mathrm{CE}}(f(\hat{\textbf{x}}^{(\ell)}), t) + \lambda_{\mathrm{mse}}\|\textbf{x} - \hat{\textbf{x}}^{(\ell)}\|_2^2\right) \right],
  \label{eq:loss-progressive}
\end{equation}
where $\ell \sim \mathrm{Uniform}\{1, \ldots, L_{\text{max}}\}$ denotes a randomly sampled trit-plane prefix, $\hat{\textbf{x}}^{(\ell)}$ is the partial reconstruction obtained from the corresponding partial bits, and $R^{(\ell)}$ is the bitrate of the trit-planes up to level $\ell$.
This ensures that the rate-accuracy trade-off is optimized not only at full bitrate but across the entire progressive decoding trajectory.

\begin{figure*}[t]
  \centering
  \begin{subfigure}[t]{0.2\textwidth}
    \centering
    \includegraphics[width=\linewidth]{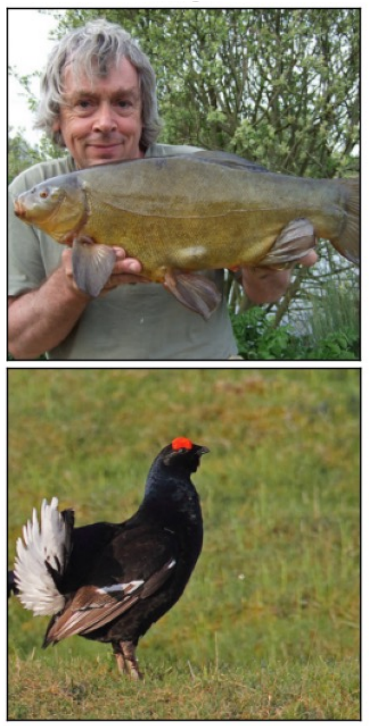}
    \caption{Original}
    \label{fig:prioritization-original}
  \end{subfigure}%
  \begin{subfigure}[t]{0.2\textwidth}
    \centering
    \includegraphics[width=\linewidth]{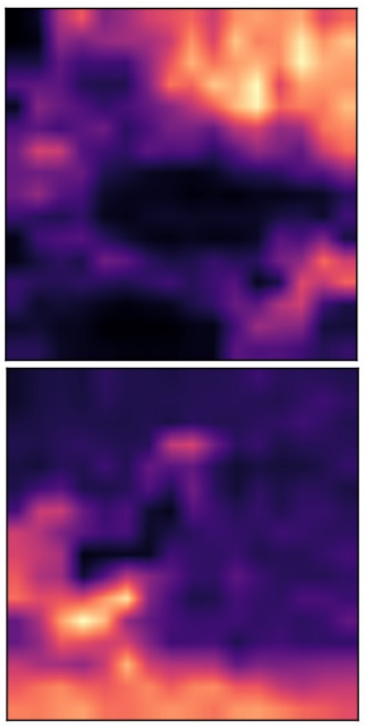}
    \caption{for Human}
    \label{fig:prioritization-human}
  \end{subfigure}%
  \begin{subfigure}[t]{0.2\textwidth}
    \centering
    \includegraphics[width=\linewidth]{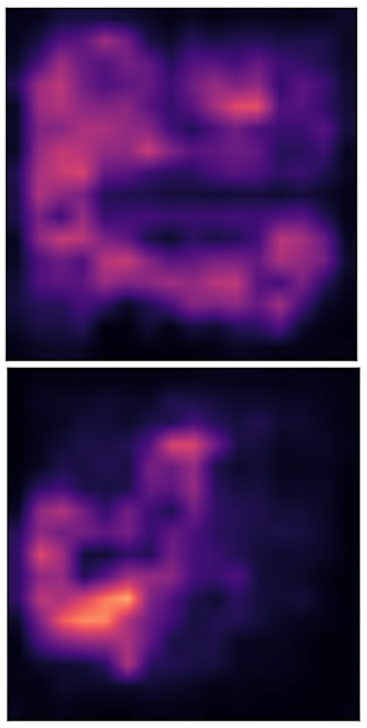}
    \caption{for Machine}
    \label{fig:prioritization-machine}
  \end{subfigure}%
  \begin{subfigure}[t]{0.2\textwidth}
    \centering
    \includegraphics[width=\linewidth]{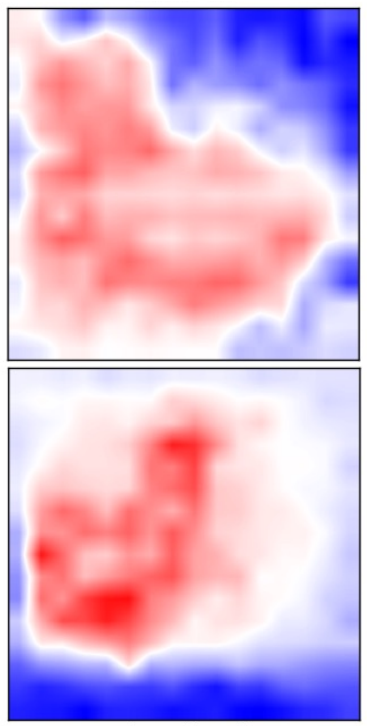}
    \caption{Difference}
    \label{fig:prioritization-difference}
  \end{subfigure}%
  \begin{subfigure}[t]{0.2\textwidth}
    \centering
    \includegraphics[width=\linewidth]{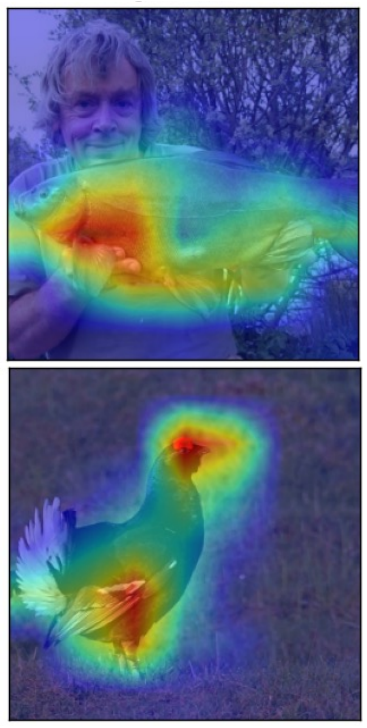}
    \caption{Grad-CAM}
    \label{fig:prioritization-gradcam}
  \end{subfigure}
  \caption{
    Symbol prioritization differences under the same bit budget (0.15 bpp) for the (b) human-oriented and (c) machine-oriented codecs. (d) The difference map shows redistribution of bits between the two codecs (red: machine allocates more, blue: less), and (e) Grad-CAM~\cite{selvaraju2017grad} visualizes the task-discriminative regions of a pretrained ResNet-50~\cite{he2016deep} on the original image.}
  \label{fig:prioritization}
  \vspace{-1.5em}
\end{figure*}

\vspace{0.5em}
\noindent \textbf{Symbol prioritization for machine perception.}
To understand how machine-oriented adaptation shifts the prioritization, we compare spatial bit maps of the human-oriented and our adapted codec under the same bit budget (see Fig.~\ref{fig:prioritization}).
For each codec, we count how many trit-planes are decoded per spatial location up to a fixed bitrate, and sum across channels to obtain a spatial allocation map representing where bits are spent.
The human-oriented codec prioritizes its budget broadly across the spatial domain (Fig.~\ref{fig:prioritization-human}), reflecting its pixel-level reconstruction objective.
In contrast, the machine-oriented codec concentrates bits on a sparser set of regions (Fig.~\ref{fig:prioritization-machine}).

The difference map (Fig.~\ref{fig:prioritization-difference}) shows that the machine-oriented codec spends additional bits (red) on the task-discriminative areas identified by Grad-CAM~\cite{selvaraju2017grad} (Fig.~\ref{fig:prioritization-gradcam}), while bits are removed from task-irrelevant background regions (blue).
This shows that machine-oriented adaptation implicitly induces task-aware symbol prioritization in the progressive bitstream, directing early-transmitted bits toward semantically critical regions without explicit supervision on the transmission order.

\subsection{Adaptive Decoding Controller}
\label{sec:methods-adaptive-decoding-controller}
  
\noindent \textbf{Motivation.}
In conventional human-oriented progressive image coding, the decoding process is typically guided by reconstruction fidelity, so later parts of the bitstream are retained as long as they improve perceptual quality.
For machine perception, however, the relevant criterion is not visual fidelity itself but whether the current reconstruction is sufficient for reliable downstream inference.
This raises a crucial question: \textbf{to what extent should we decode a progressive bitstream to ensure sufficient suitability of the current reconstruction for the downstream machine prediction while minimizing unnecessary bit transmission?}
Our adaptive decoding controller addresses this by dynamically selecting the minimum decoding level sufficient for reliable downstream inference.

\begin{figure}[t]
\begin{minipage}[t]{0.485\linewidth}
\begin{algorithm}[H]
\caption{Training Adaptive Decoding Controller}
\label{alg:training}
\scriptsize
\begin{algorithmic}[1]
\Require Dataset $\mathcal{D} = \{(\textbf{x}_i, t_i)\}_{i=1}^N$, Classifier $f$, Levels $\mathcal{L} = \{1, \ldots, L_{\text{max}}\}$
\Ensure Trained controller $\mathcal{C}$
\State Initialize empty feature vector $\mathbf{X}$ and label vector $\mathbf{s}$
\For{each $(\textbf{x}, t)$ in $\mathcal{D}$}
    \For{each level $\ell \in \mathcal{L}$}
        \State $\hat{\textbf{x}}^{(\ell)} \leftarrow \textsc{Compress}(\textbf{x}, \ell)$
        \State $\textbf{v}^{(\ell)}  \leftarrow f(\hat{\textbf{x}}^{(\ell)})$
        \State $\hat{t}^{(\ell)}  \leftarrow \arg\max_k v_k^{(\ell)} $
        \State $\boldsymbol{\phi}^{(\ell)}  \leftarrow \textsc{ComputeFeature}(\textbf{z}^{(\ell)} )$
        \State $s^{(\ell)} \leftarrow \mathds{1}[\hat{t}^{(\ell)} = t]$
        \State Append $\boldsymbol{\phi}^{(\ell)}$ to $\mathbf{X}$
        \State Append $s^{(\ell)}$ to $\mathbf{s}$
    \EndFor
\EndFor
\State Fit model $\mathcal{C}$ on $(\mathbf{X}, \mathbf{s})$ with logistic regression
\State \Return $\mathcal{C}$
\end{algorithmic}
\end{algorithm}
\end{minipage}
\hfill
\begin{minipage}[t]{0.485\linewidth}
\begin{algorithm}[H]
\caption{Task-aware Progressive Decoding with Adaptive Decoding Controller}
\label{alg:inference}
\scriptsize
\begin{algorithmic}[1]
\Require Bitstream $\mathcal{B}$, Decoder $g_s(\mathcal{B}, \ell)$, Classifier $f$, Controller $\mathcal{C}$
\Ensure Predicted label $\hat{y}$ and chosen level $\ell_{\tau}$
\For{$\ell = 1$ to $L_{\text{max}}$}
    \State $\hat{\textbf{x}}^{(\ell)} \leftarrow g_s(\mathcal{B}, \ell)$
    \State $\textbf{v}^{(\ell)} \leftarrow f(\hat{\textbf{x}}^{(\ell)})$
    \State $\hat{t}^{(\ell)} \leftarrow \arg\max_k v_k^{(\ell)}$
    \State $\boldsymbol{\phi}^{(\ell)} \leftarrow \textsc{ComputeFeature}(\textbf{z}^{(\ell)})$
    \State $\mathcal{S}^{(\ell)} \leftarrow \mathcal{C}(\boldsymbol{\phi}^{(\ell)})$
    \If{$\mathcal{S}^{(\ell)} \geq \tau$}
        \State \Return $\hat{t}^{(\ell)}, \ell_{\tau} \leftarrow \ell$
    \EndIf
\EndFor
\State \Return $\hat{t}^{(L_{\text{max}})},\, \ell_{\tau} \leftarrow L_{\text{max}}$
\end{algorithmic}
\end{algorithm}
\end{minipage}
\vspace{-1.5em}
\end{figure}

\vspace{0.5em}
\noindent \textbf{Systematic design.}
Our adaptive decoding controller determines the minimum bits needed to maintain the desired confidence of downstream machine prediction for each input image.
Given a progressive codec that produces reconstructions $\{\hat{\textbf{x}}^{(1)}, \ldots, \hat{\textbf{x}}^{(L_{\text{max}})}\}$ from low-bitrate to high-bitrate, our goal is to select the smallest level $\ell_{\tau}$ such that the downstream machine (e.g., classifier $f$) achieves acceptable confidence ($\tau$).
The key idea is to train a lightweight controller on the classifier's output statistics--logit-space and softmax-based confidence signals--to predict whether the current reconstruction suffices for reliable inference.

To train the controller $\mathcal{C}$, we compress images from the ImageNet train set~\cite{russakovsky2015imagenet} at each progressive decoding level $\ell \in \{1, \ldots, L_{\text{max}}\}$ and collect the corresponding reconstructions $\hat{\textbf{x}}^{(\ell)}$ (see Algorithm~\ref{alg:training}).


We adopt a simple logistic regression model that operates on a 12-dimensional feature vector $\boldsymbol{\phi}$ extracted from the classifier's output at each decoding level, following the design of~\cite{pouget2025suitability}: softmax-based statistics computed from the probability vector $\textbf{p}$ ($p_1$, $\mathrm{std}(\textbf{p})$, entropy, $p_1/p_2$, and the cumulative top-10 probability), logit-based statistics computed from the logit vector $\textbf{v}$ ($\mathrm{mean}(\textbf{v})$, $v_1$, $\mathrm{std}(\textbf{v})$, and $v_1{-}v_2$), and energy-based statistics ($-\log p_1$, $\log(p_1/p_2)$, and $-\mathrm{logsumexp}(\textbf{v})$).


For each sample, we assign a binary label $s = \mathds{1}[\hat{t} = t]$ indicating whether the classifier's prediction is correct, and train $\mathcal{C}(\boldsymbol{\phi}) = \Pr(s=1 \mid \boldsymbol{\phi})$ to estimate the expected correctness of the downstream prediction--which we refer to as the suitability of the current reconstruction--at a given decoding level.
This enables the controller to identify the smallest level that satisfies the target confidence threshold $\tau$.

At inference time, we decode progressively from $\ell=1$ (see Algorithm~\ref{alg:inference}).
At each level, we obtain the estimated suitability $\mathcal{S}^{(\ell)} = \mathcal{C}(\boldsymbol{\phi}^{(\ell)})$.
If $\mathcal{S}^{(\ell)} \geq \tau$ for a user-specified threshold $\tau$, we stop decoding.
Otherwise, we proceed to $\ell+1$.
This enables machine-adaptive bit allocation driven by predicted suitability, rather than fixed quality targets.

\section{Experiments}
\label{sec:experiments}

\subsection{Training}
\label{sec:experiments-training}

\noindent\textbf{Codec backbone.}
We adopt a pretrained TransTIC base codec\footnote{\small\url{https://github.com/NYCU-MAPL/TransTIC/base_codec_4.pth.tar}} as the frozen human codec backbone, which follows the TIC architecture~\cite{lu2022tic} with a parallel mean-scale hyperprior ($M{=}192$ latent channels, $N{=}128$ hyperprior channels).
The backbone is kept fully frozen throughout all training stages.

\vspace{0.5em}
\noindent\textbf{Training dataset.}
We sample 80,000 images from the ImageNet-1K~\cite{russakovsky2015imagenet} training set for adaptation module training.

\vspace{0.5em}
\noindent\textbf{Machine adaptation.}
We train only the lightweight adaptation modules: three SFMA modules in each of the encoder and decoder, and one HSLoRA module in the hyperprior decoder (rank$=16$).
The trainable parameters total 300,080, which correspond to 3.84\% of the full network.
We optimize them with the objective defined in Eq.~\ref{eq:loss-function}, setting $\lambda_{\mathrm{task}}{=}0.8$ and $\lambda_{\mathrm{mse}}{=}0.0025$, with a frozen ResNet-50\footnote{\url{https://huggingface.co/timm/resnet50.a1_in1k}}~\cite{he2016deep} as the downstream classifier $f$.
Progressive-decoding aware training is also applied as described in Sec.~\ref{sec:method-adaptation}.

\vspace{0.5em}
\noindent\textbf{Optimization.}
We train for 60 epochs using AdamW~\cite{loshchilov2018decoupled} with learning rate $10^{-4}$, weight decay $10^{-4}$, and cosine annealing ($\eta_{\min}{=}10^{-6}$).

\vspace{0.5em}
\noindent\textbf{Adaptive decoding controller.}
Following Sec.~\ref{sec:methods-adaptive-decoding-controller}, we separately compress 5,000 images from the ImageNet-1K training set at each progressive decoding level and collect the corresponding classifier (ResNet-50) output features.
A logistic regression model is then fitted on the resulting feature-label pairs to predict inference correctness at each decoding level.
For ConvNeXt\footnote{\url{https://huggingface.co/timm/convnext_base.fb_in1k}}~\cite{liu2022convnet}, we additionally train a separate adaptive decoding controller.

\subsection{Evaluation}
We follow the settings in~\cite{li2024image} to evaluate the performance of our method and baselines.
The evaluation is done on the validation set of ImageNet-1K~\cite{russakovsky2015imagenet}.
Images are resized to 256 $\times$ 256 for compression, and center cropped to 224 $\times$ 224 with normalization for evaluation.
We use the top-1 accuracy as a performance metric.
We evaluate the performance using ResNet-50~\cite{he2016deep} and ConvNeXt~\cite{liu2022convnet} from the timm~\cite{rw2019timm} library.
All experiments are conducted on NVIDIA A100-SXM4-40GB GPUs.

\begin{figure*}[t]
  \centering
  \begin{subfigure}[t]{0.42\textwidth}
    \centering
    \includegraphics[width=\linewidth]{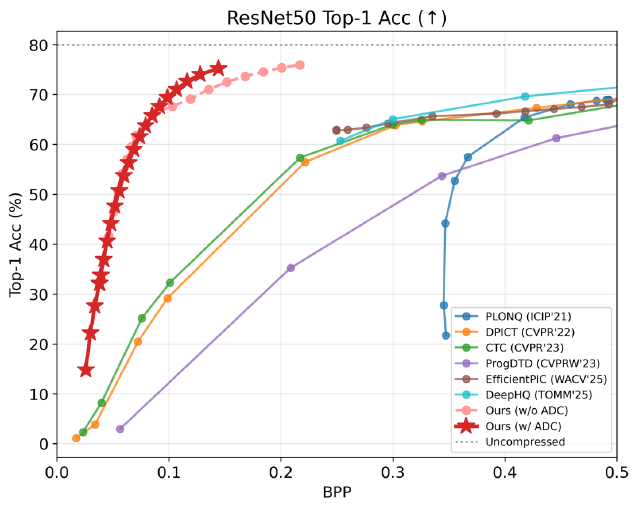}
    \caption{Progressive codecs}
    \label{fig:rd-resnet-progressive}
  \end{subfigure}%
  \hspace{0.5em}
  \begin{subfigure}[t]{0.42\textwidth}
    \centering
    \includegraphics[width=\linewidth]{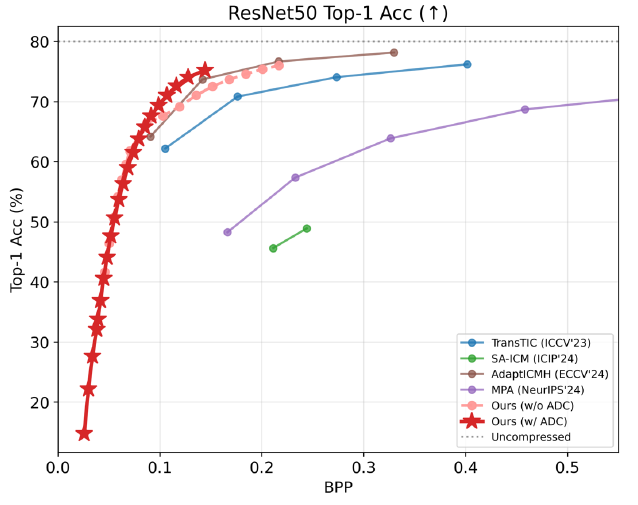}
    \caption{Machine-oriented codecs}
    \label{fig:rd-resnet-machine}
  \end{subfigure}
  \begin{subfigure}[t]{0.42\textwidth}
    \centering
    \includegraphics[width=\linewidth]{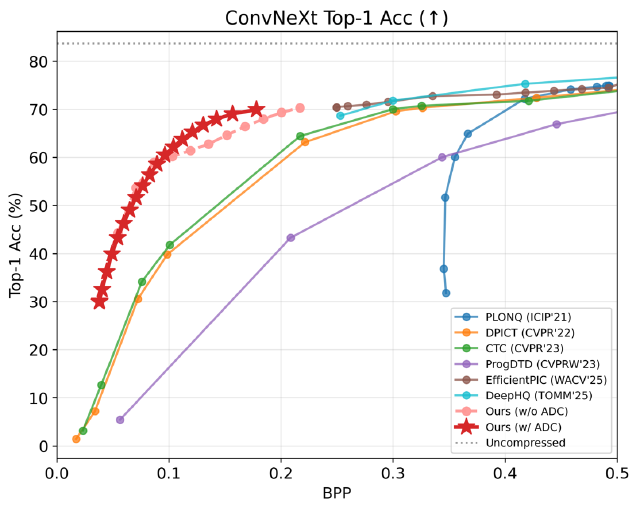}
    \caption{Progressive codecs}
    \label{fig:rd-convnext-progressive}
  \end{subfigure}%
  \hspace{0.5em}
  \begin{subfigure}[t]{0.42\textwidth}
    \centering
    \includegraphics[width=\linewidth]{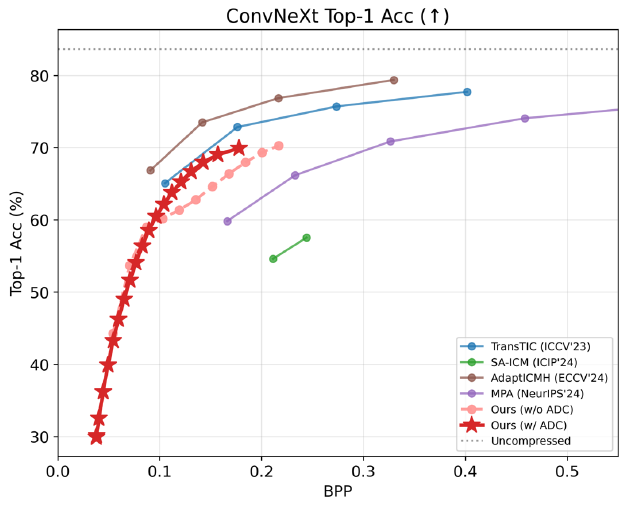}
    \caption{Machine-oriented codecs}
    \label{fig:rd-convnext-machine}
  \end{subfigure}
  \vspace{-0.5em}
  \caption{
    Rate-accuracy performance comparison. The upper panels (a, b) show results on ResNet-50~\cite{he2016deep}, while the lower panels (c, d) show results on ConvNeXt~\cite{liu2022convnet}.
    The left panels (a, c) compare PICM-Net against progressive human-oriented codecs, while the right panels (b, d) compare against machine-oriented non-progressive codecs. In each panel, the pink circles denote PICM-Net without the adaptive decoding controller, and the red stars denote PICM-Net with the controller. Gray dashed lines indicate reference performance on uncompressed images.}
  \label{fig:rd-comparison}
  \vspace{-1.5em}
\end{figure*}

\subsection{Baselines}
To demonstrate the effectiveness of our proposed framework, we compare our codec, PICM-Net, with the state-of-the-art learned image codecs.
Since there is no existing machine-oriented progressive learned image codec, we indirectly compare performance with two groups: human-oriented progressive codecs and machine-oriented non-progressive codecs.
For the former, we adopt state-of-the-art human-oriented progressive learned image codecs: PLONQ~\cite{lu2021progressive}, DPICT~\cite{lee2022dpict}, CTC~\cite{jeon2023context}, ProgDTD~\cite{hojjat2023progdtd}, Efficient-PIC~\cite{presta2025efficient}, and DeepHQ~\cite{lee2024deephq}.
For the latter, we adopt four machine-oriented non-progressive codecs: TransTIC~\cite{chen2023transtic}, AdaptICMH~\cite{li2024image}, SA-ICM~\cite{shindo2024image}, and MPA~\cite{zhang2024all}.
All codecs are implemented based on the CompressAI~\cite{begaint2020compressai} library.

\begin{figure*}[t]
  \centering
  \begin{subfigure}[t]{0.19\textwidth}
    \centering
    \includegraphics[width=\linewidth]{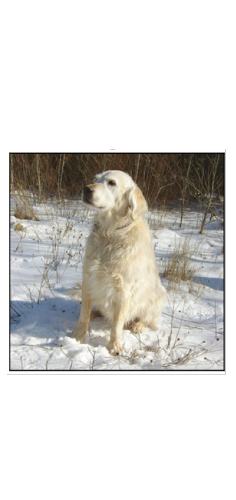}
    \caption{Original}
    \label{fig:qualitative-original}
  \end{subfigure}%
  \hfill
  \begin{subfigure}[t]{0.19\textwidth}
    \centering
    \includegraphics[width=\linewidth]{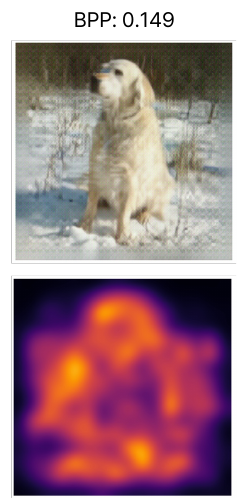}
    \caption{PICM-Net}
    \label{fig:qualitative-ours}
  \end{subfigure}%
  \hfill
  \begin{subfigure}[t]{0.19\textwidth}
    \centering
    \includegraphics[width=\linewidth]{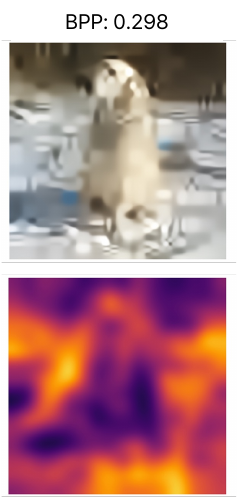}
    \caption{PLONQ~\cite{lu2021progressive}}
    \label{fig:qualitative-plonq}
  \end{subfigure}%
  \hfill
  \begin{subfigure}[t]{0.19\textwidth}
    \centering
    \includegraphics[width=\linewidth]{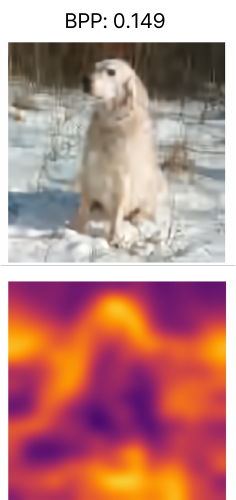}
    \caption{DPICT~\cite{lee2022dpict}}
    \label{fig:qualitative-dpict}
  \end{subfigure}%
  \hfill
  \begin{subfigure}[t]{0.19\textwidth}
    \centering
    \includegraphics[width=\linewidth]{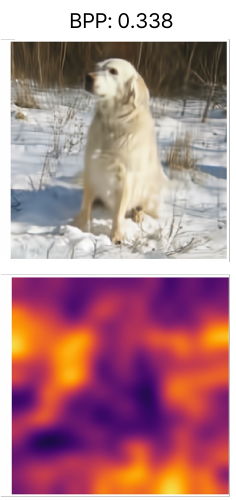}
    \caption{DeepHQ~\cite{lee2024deephq}}
    \label{fig:qualitative-deephq}
  \end{subfigure}
  \vspace{-0.5em}
  \caption{
    Qualitative comparison of reconstructed images and their bit allocation maps from PICM-Net and representative progressive baseline codecs. All methods are compared at similar bitrate levels to highlight prioritization differences under progressive decoding. Each bit allocation map is normalized to represent its prioritization.}
  \label{fig:qualitative}
  \vspace{-1.5em}
\end{figure*}

\subsection{Rate-Accuracy Performance}

\noindent\textbf{Comparison with progressive codecs.}
Fig.~\ref{fig:rd-resnet-progressive} and~\ref{fig:rd-convnext-progressive} compare PICM-Net against human-oriented progressive codecs on ResNet-50~\cite{he2016deep} and ConvNeXt~\cite{liu2022convnet}.
Across both classifiers, PICM-Net consistently achieves superior rate-accuracy performance across all bitrates, outperforming all baselines.
This gain is directly attributable to machine-oriented adaptation: unlike existing progressive codecs optimized for perceptual quality, PICM-Net prioritizes task-discriminative information in early bits.

\vspace{0.5em}
\noindent\textbf{Comparison with machine-oriented codecs.}
Fig.~\ref{fig:rd-resnet-machine} and~\ref{fig:rd-convnext-machine} compare PICM-Net against machine-oriented non-progressive codecs.
Since existing machine-oriented codecs are each optimized at fixed discrete bitrate points, a direct rate-accuracy comparison is inherently asymmetric.
Nevertheless, PICM-Net achieves competitive rate-accuracy performance on ResNet-50, while uniquely supporting fine-grained progressive decoding from a single bitstream.
On ConvNeXt, a moderate performance gap is observed, which we attribute to the adaptation modules being trained with ResNet-50 as the downstream classifier, thus generalization to unseen classifiers remains a direction for future work.

\vspace{0.5em}
\noindent\textbf{Qualitative results.}
Fig.~\ref{fig:qualitative} shows progressively reconstructed images and the corresponding bit allocation maps.
In particular, representative progressive codecs (PLONQ~\cite{lu2021progressive}, DPICT~\cite{lee2022dpict}, and DeepHQ~\cite{lee2024deephq}) allocate more bits to background regions with large color variations rather than to the foreground object, whereas PICM-Net preferentially restores the object region.

\subsection{How Adaptive Decoding Controller Contributes}

As shown in Fig.~\ref{fig:rd-comparison}, PICM-Net with and without the adaptive decoding controller exhibit nearly identical rate-accuracy at low bitrates, while the controller yields substantially better performance at higher bitrates.
This pattern is intuitive: at low bitrates, most images have not yet reached a reliable downstream prediction, so there is little redundancy for the controller to remove.
At higher bitrates, however, many images are already correctly classified well before the full bitstream is decoded--additional bits improve reconstruction fidelity but no longer change the downstream decision.
The controller exploits this property by stopping decoding per image as soon as the predicted suitability exceeds the threshold $\tau$.

\vspace{0.5em}
\noindent\textbf{Reallocation of bits from easy to hard images.}
To examine whether the controller performs meaningful per-sample bit reallocation, we compare it against a matched static decoder fixed at a single decoding level chosen to match the mean bitrate at $\tau{=}0.70$.
For each image $i$, we define the per-sample rate difference $\Delta r_i = r_i^{\mathrm{ctrl}} - r_i^{\mathrm{static}}$, where $\Delta r_i < 0$ indicates the controller saves bits and $\Delta r_i > 0$ indicates it spends additional bits.

\begin{figure}[t]
\centering
\begin{subfigure}[t]{0.38\linewidth}
  \includegraphics[width=\linewidth]{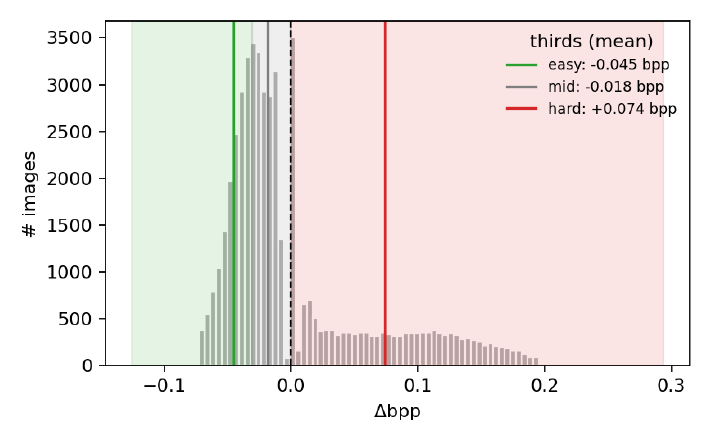}
  \caption{Per-image $\Delta r$ distribution.}
  \label{fig:adc-hist}
\end{subfigure}
\begin{subfigure}[t]{0.61\linewidth}
  \includegraphics[width=\linewidth]{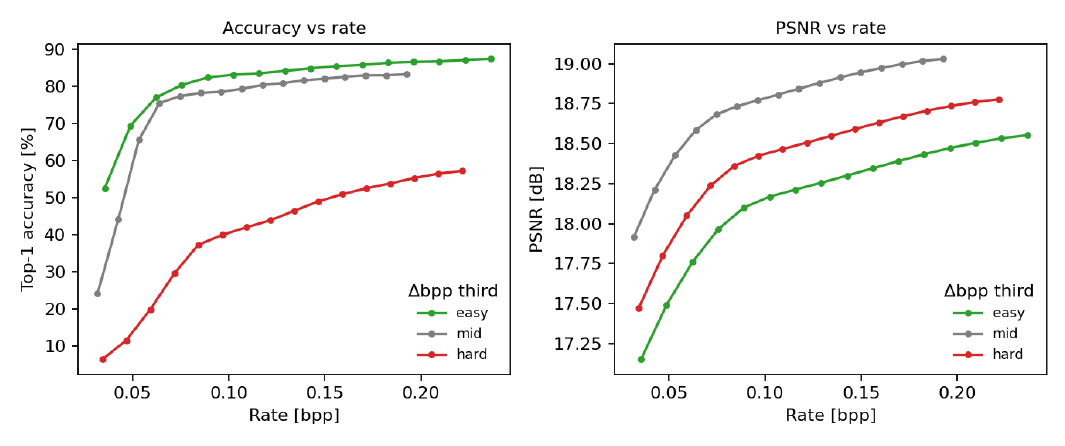}
  \caption{Accuracy and PSNR by $\Delta r$ group.}
  \label{fig:adc-thirds-curves}
\end{subfigure}
\vspace{-0.5em}
\caption{
  Adaptive decoding controller analysis on 50K ImageNet-1K validation images with ResNet-50 ($\tau{=}0.70$).
  (a) The controller saves bits on a large subset of images while spending more on others, resulting in a skewed distribution centered below zero.
  (b) Unlike task accuracy, which saturates early for rate-saved images (green), PSNR continues to rise monotonically with bitrate across all groups--confirming that bits removed by the controller improve reconstruction fidelity but do not contribute to the downstream decision.
}
\label{fig:adc-analysis}
\vspace{-1.5em}
\end{figure}

\begin{figure}[t]
\centering
\includegraphics[width=0.9\linewidth]{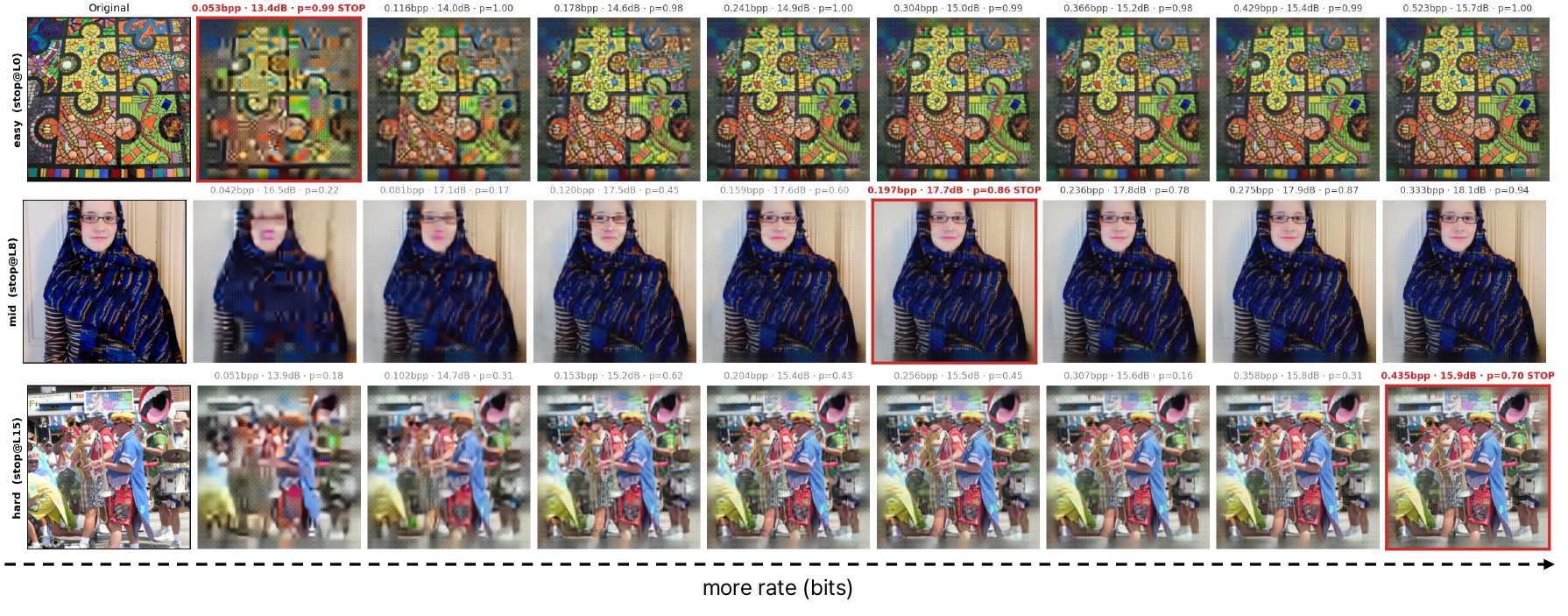}
\vspace{-1.0em}
\caption{
  Progressive decoding strips for representative easy (top), mid (middle), and hard (bottom) images.
  Red-bordered frames indicate the stopping point of the adaptive decoding controller ($\tau{=}0.70$).
  For the easy image, the classifier reaches a correct prediction at an early decoding level (0.053 bpp), where the estimated suitability already exceeds the desired confidence and subsequent decoding only improves PSNR.
  For the hard image, the suitability does not exceed the desired confidence level until near the final decoding level (0.435 bpp), where additional bits are genuinely necessary for reliable inference.
}
\label{fig:adc-strips}
\vspace{-1.5em}
\end{figure}

Fig.~\ref{fig:adc-hist} shows that the controller does not uniformly reduce rate: it saves bits for a large subset of images while spending more on others.
Partitioning images into three equal groups by $\Delta r_i$, the rate-saved group (easy group in Fig.~\ref{fig:adc-analysis}) achieves 87.4\% full-rate accuracy--indicating these are relatively easy samples--while the rate-spent group (hard group in Fig.~\ref{fig:adc-analysis}) reaches only 57.2\%, suggesting these are harder samples that benefit from additional decoding.
Accordingly, the rate-saved group stops after decoding roughly the first 4\% of the progressive trit-plane trajectory on average, whereas the rate-spent group continues to roughly 67\% on average.
Fig.~\ref{fig:adc-thirds-curves} further confirms this: for the rate-saved group, accuracy saturates early and additional decoding yields diminishing returns, whereas the rate-spent group remains rate-sensitive throughout--while PSNR keeps rising for all groups regardless of difficulty.

Fig.~\ref{fig:adc-strips} illustrates this behavior at the sample level across easy, mid, and hard examples.
For the easy image, the classifier reaches a correct prediction (the estimated suitability exceeds the desired confidence $\tau$) at an early decoding level, after which PSNR continues to improve but the predicted class and suitability remain stable.
For the hard image, the controller output stays below $\tau$ until a much later decoding level, demonstrating that the controller naturally extends decoding where additional bits are genuinely needed.

\vspace{0.5em}
\noindent\textbf{Quantitative gains.}
In the practical operating region (accuracy $\geq 70\%$), the controller achieves a BD-rate of $-21.6\%$ relative to the decoder without our controller.
At fixed desired confidences ($\tau$) of 70\%, 73\%, and 75\%, the average bitrate is reduced by 18.7\%, 23.9\%, and 26.4\%, respectively.
Outside this operating region, the benefit of the controller is limited: at very low bitrates, the fixed threshold may trigger premature stopping, resulting in an accuracy decrease of up to $1.91\text{\%p}$ relative to static decoding.

\subsection{Ablation Studies}
\label{sec:ablation}

\noindent\textbf{Component Ablation.}
We measure the contribution of each component in terms of BD-rate relative to the frozen backbone.
Adding SFMA alone yields the largest gain ($-66.9\%$ BD-rate in the operating region), confirming that spatial-frequency modulation of the encoder and decoder features mainly contributes to machine-oriented adaptation.
HSLoRA further reduces BD-rate by 1.9\%, realigning the entropy model to the adapted latent distribution, without changing the accuracy achieved at the full decoding level ($\ell=L_{\max}$).
Removing progressive-decoding aware training causes the rate-accuracy curve to collapse across all bitrates: the average accuracy over the progressive bitrates falls from $66.5\%$ to $54.1\%$, demonstrating that progressive-decoding aware training is essential for optimizing performance across all decoding bitrates.

\vspace{0.5em}
\noindent \textbf{Prioritization strategy.}
We compare sigma-based and expected variance-based symbol ordering.
Expected variance-based ordering achieves a marginal BD-rate gain of $-0.4\%$, but incurs $20.7\times$ slower symbol-order prioritization and $2.28\times$ slower overall decoding time (21.0 vs.\ 9.2 ms per image).
We therefore adopt sigma-based ordering, which achieves comparable task performance at substantially lower computational cost.

\section{Conclusion}
\label{sec:conclusion}
In this paper, we presented PICM-Net, the first progressive learned image codec for machine perception, combining progressive trit-plane coding, machine-oriented adaptation (SFMA and HSLoRA) with progressive-decoding aware training, and an adaptive decoding controller that determines the minimum decoding level sufficient for reliable inference on a per-image basis.
Extensive experiments showed that PICM-Net achieves superior rate-accuracy performance against progressive codecs and competitive performance against machine-oriented non-progressive codecs, while uniquely enabling progressive decoding from a single bitstream.

\vspace{0.5em}
\noindent\textbf{Future work.}
Since the adaptation modules are currently trained with a specific downstream classifier, a promising direction is to make the adaptation classifier-agnostic and to extend our framework beyond classification to dense machine tasks such as object detection and segmentation.

\newpage
\section*{Acknowledgements}
This work was supported by the National Research Foundation of Korea (NRF) grant (No. RS-2024-00453301, RS2025-00517159) and by Institute of Information \& communications Technology Planning \& Evaluation (IITP) grant (IITP-2025-RS-2024-00428780).

\bibliographystyle{splncs04}
\bibliography{main}



\end{document}